\documentclass[10pt,journal,compsoc]{IEEEtran}

\usepackage{float}
\usepackage{tabu}
\usepackage{color}
\usepackage{hyperref}
\usepackage{amsmath}
\usepackage{amsfonts}
\usepackage{amssymb}
\usepackage{graphicx}
\usepackage{booktabs}
\usepackage{algorithm}
\usepackage[noend]{algpseudocode}
\usepackage{comment}
\newtheorem{definition}{Definition}
\usepackage[
singlelinecheck=false 
]{caption}

\ifCLASSOPTIONcompsoc
  \usepackage[nocompress]{cite}
\else
  \usepackage{cite}
\fi
\ifCLASSINFOpdf
\else
\fi
\begin{document}
%
\title{Metadata-Based Detection of Child Sexual Abuse Material}
%
%
%
%

\author{Mayana~Pereira,
        Rahul~Dodhia,
        Hyrum~Anderson,
        and~Richard~Brown
\IEEEcompsocitemizethanks{\IEEEcompsocthanksitem M. Pereira is with Microsoft Corporation, AI for Good Research Lab, Redmond, WA, 98052 and also with the Universidade de Bras\' ilia, Campus Darcy Ribeiro, Bras\' ilia, Brazil.\protect\\
E-mail: mayana.wanderley@microsoft.com

\IEEEcompsocthanksitem R. Dodhia and H. Anderson are with Microsoft Corporation, Redmond,
WA, 98052.\protect\\
\IEEEcompsocthanksitem R. Brown is with Project VIC International.}
}

\IEEEtitleabstractindextext{%
\begin{abstract}
Child Sexual Abuse Media (CSAM) is defined as any visual record of a sexually explicit activity involving minors. CSAM impacts victims differently from the actual abuse because the distribution never ends, and images are permanent. Machine learning-based solutions can help law enforcement quickly identify CSAM and even block its digital distribution. However, collecting and storing CSAM imagery to train machine learning models has many ethical and legal constraints, creating a barrier to research development in this space. With such restrictions in place, the development of CSAM machine learning detection systems based on file metadata opens up several opportunities. File metadata is not a record of a crime, and it does not have legal restrictions. Therefore, investing efforts in developing detection systems based on metadata can increase the rate of discovery of CSAM and potentially help thousands of victims. 
 
 Our contributions are two-fold: First, we propose a framework for training and evaluating deployment-ready machine learning models for CSAM identification.  Our framework provides guidelines on how to evaluate a CSAM detection model against intelligent adversaries and guidelines on how to assess CSAM detection models with open data. We than apply the proposed framework to the problem of CSAM detection on file storage systems based solely on metadata - namely, file paths. The resulting model is a media-agnostic tool (image, video, PDF) that can potentially scan thousands of file storage systems in a short time. In our experiments, the best-performing model is based on convolutional neural networks (CNN) and achieves an accuracy of 0.97 and recall of 0.94. Our testing framework provides concrete evidence that our CNN-based model is robust against offenders actively trying to evade detection by evaluating the model against adversarial modifications in the file paths. Additionally, experiments with open datasets prove that the model generalizes well and is deployment-ready, achieving a false positive rate of 0.005 in data samples from distinct data distributions.
 
 \end{abstract}

}

\maketitle

\IEEEdisplaynontitleabstractindextext

%
\IEEEpeerreviewmaketitle

\IEEEraisesectionheading{\section{Introduction}\label{sec:introduction}}

International law enforcement handles millions of child sexual abuse cases annually. In 2017, child abuse hotlines received and reviewed over 37 million child sexual abuse files worldwide \cite{vic}. Despite the 2008 \textit{Protect our children act} \cite{protect}, the quantity of CSAM in digital platforms has dramatically grown over the last decade\cite{wolak2014measuring}. Online sharing platforms and social media facilitated\cite{10.1093/police/pay028} the explosive growth of CSAM creation and distribution \cite{ncmec_google}. Every platform for content searching and sharing, including social material, likely has CSAM on it \cite{keller_dance_2019}. Recent studies have reported alarming statistics, 25\% of girls and 17\% of boys in the U.S. have experienced some form of sexual abuse before the age of 18 \cite{unicef}.
As the scale of the problem grows, technology plays an essential role in CSAM identification. Companies that manage user-generated data, such as Pinterest, Facebook, Microsoft, Apple, and Google, have made detection and removal of CSAM a top priority. Although several non-profit organizations such as Project VIC International\footnote{https://www.projectvic.org}, Thorn\footnote{https://www.thorn.org} and the Internet Watch Foundation\footnote{https://www.iwf.org.uk} focus on building tools to combat CSAM proliferation, the creation and distribution of CSAM is still a growing problem.

The COVID-19 pandemic triggered a significant increase in the distribution of CSAM via social media and video conferencing apps \cite{olivia.2020}. The identification of CSAM is a highly challenging problem. First, it can manifest in different types of material: images, videos, streaming, video conference, online gaming, among others. Undiscovered and unlabeled CSAM on the internet is estimated to be magnitudes greater than the currently identified CSAM. Second, discovering new material is still highly dependent on human discovery. Despite the significant progress in machine learning models for CSAM identification with modern deep-learning architectures \cite{vitorino2016two, macedo2018benchmark, research1}, these models rely on the availability of labeled images, which can lead to technical limitations. As new material is created daily, we understand that utilizing complementary signals can advance the capability of digital platforms in detecting and removing illegal content. The use of metadata has been proposed in the past by \cite{peersman2016icop}. This is an effective approach since distributors use coded language to communicate and trade links of CSAM hosted in plain sight on content sharing platforms such as Facebook, YouTube, Twitter, and Instagram.  

The scarcity of frameworks for training and evaluation deployment-ready machine learning models for CSAM detection prevents a broader adoption of machine learning pipelines for CSAM detection. Before deployment, organizations should test the CSAM detection model under different conditions. An evaluation scenario needs a real-world dataset with similar data distributions to what the model will get exposed to after deployment. A critical scenario for analysis is testing the model on completely benign out-of-sample datasets. The burden caused by a high false-positive rate can result in halting the deployment of such systems. Furthermore, it is crucial to understand how adversarially modified data samples impacts model performance. 

\subsection{Our Contributions}
\begin{figure*}
\centering
\includegraphics[width=7in]{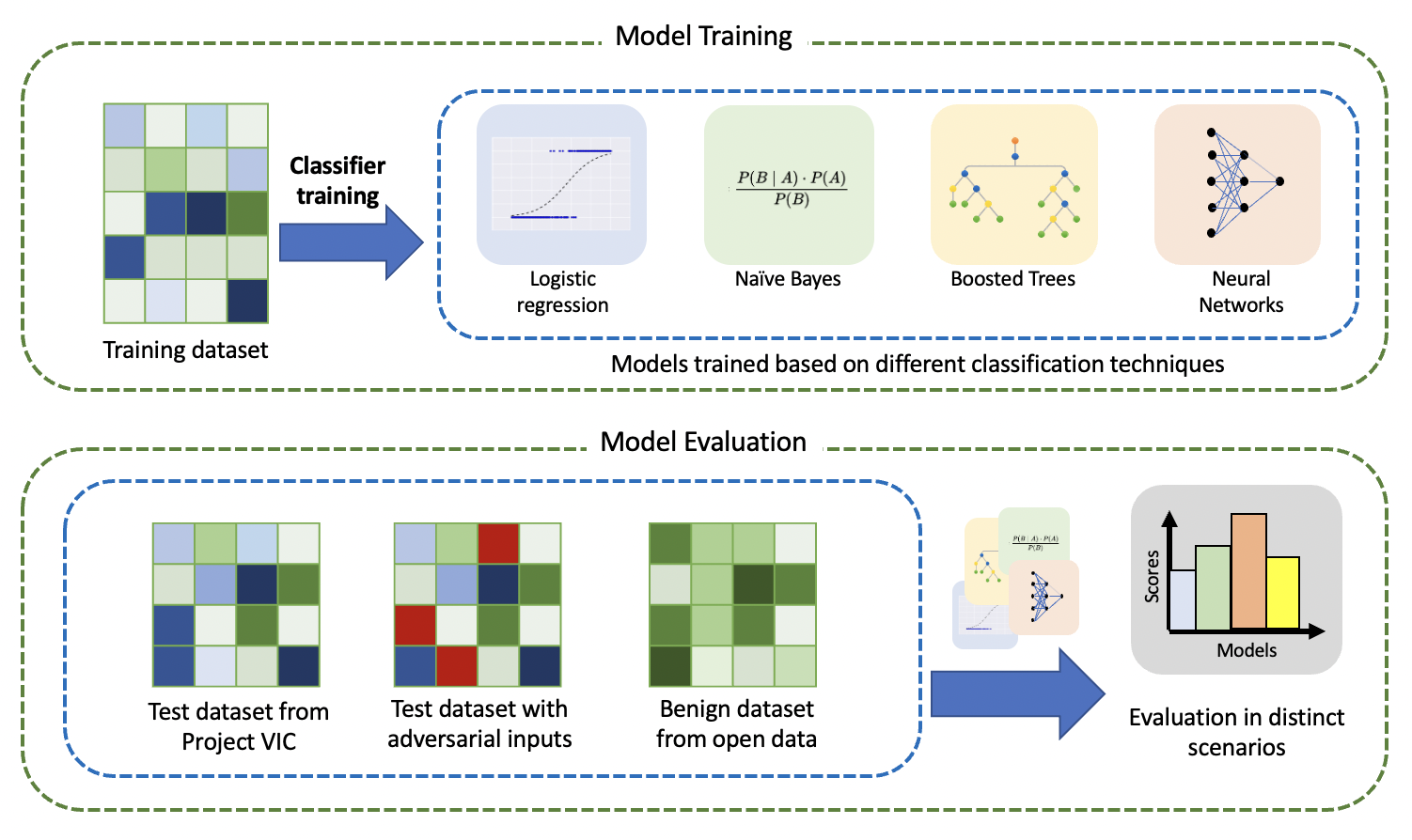}
\caption{Pipeline for model training and evaluation of machine learning models for CSAM detection. (i) During model training, we train models utilizing several machine learning techniques, such as logistic regression, Naive Bayes, boosted trees, and Neural Networks. (ii) We construct different testing datasets to model performance in different circumstances of practical relevance during the model evaluation. We propose a testing framework where the model is tested under three scenarios: File paths from out-of-sample hard drives,  file paths intentionally modified by an adversary to evade detection, and file paths from benign sources (open data).}
\label{intro}
\end{figure*}

Building machine learning systems for detecting CSAM media is a complex task. Due to the associated legal constraints, systems that relies on metadata for detecting and blocking the distribution of CSAM can expedite the hard work of NGOs and content moderators.  
This work proposes a framework for training and evaluation of deployment-ready machine learning models for CSAM detection based solely on file metadata. The proposed models compute the likelihood that a file path is associated with CSAM. Our experiments show that the resulting model not only does not rely on the availability of CSAM images and videos for model training but also achieves desired performance in challenging real-world deployment scenarios. We list the contributions as follow:
\begin{itemize} 

\item We propose a training and evaluation framework to develop deployment-ready machine learning models for CSAM text classification tasks, including file path classification, website keyword classification, search terms classification and others. Our framework includes a model training pipeline that covers several text representation and machine learning techniques. Our testing pipeline, illustrated in Figure \ref{intro}, covers real-world scenarios that should be expected when deploying a machine learning model for CSAM detection: (i) test on out-of-sample csam and non-csam samples; (ii) test on adversarially modified CSAM samples to evade detection; (iii) test on benign samples from open data sources. 

 \item We train and compare several machine-learning models that analyze file paths and file names from file storage systems and determine a probability that a given file has child sexual abuse content. We train our models on a real-world dataset containing over one million file paths. It is the most extensive file path dataset ever used for CSAM detection to date. Our best classifier achieves recall rates over 0.94 and accuracy over 0.97 on holdout sets; it maintains a high recall rate in adversarially modified inputs; when tested against benign samples from other data distributions, it achieves a false-positive rate of $\approx$ 0.01.
\end{itemize}

To our knowledge, our work is the first to propose a framework for training and evaluation of deployment-ready CSAM detection systems that include adversarial examples in the evaluation stage. Our results show that machine learning based on file paths can effectively detect CSAM in storage systems and achieve the aspired performance in all the proposed evaluation scenarios.

\section{Related Work}
 Identification of CSAM via statistical algorithms is a reasonably recent approach. In the early 2000s, the US and the UK introduced laws targeting online exploitation of minors (COPA in the US, Crime and Disorder Act UK)\cite{davidson_gottschalk}. However, only in 2008, the first widely used technology for CSAM identification was released.
 
\textbf{PhotoDNA Hash }PhotoDNA Hash (PDNA) is a widely used technique for automated identification of CSAM. The PDNA uses a  fuzzy hash algorithm to convert a CSAM image to a long string of characters. The converted hashes are compared against other hashes to find identical or similar images. PDNA technology enabled a faster discovery of CSAM while protecting the victim's identity. This system is still one of the most widely used methods for detecting CSAM images worldwide. Search engines, social networks, and image sharing services utilize databases of hashed CSAM images to eradicate harmful content from their platforms.  PDNA is a signature-based technology; it recalls only known CSAM. Therefore, identifying new CSAM in a PDNA-based system requires manual labeling. 

\textbf{Machine Learning for Image Identification } Since PDNA's first development, computer vision models have undergone a revolution resulting in novel machine learning-based models for pornography and CSAM detection \cite{nian2016pornographic, macedo2018benchmark, research1, peersman2014icop}. The current approaches either combine a computer vision model to extract image descriptors \cite{vitorino2016two}, train computer vision models on pornography data \cite{gangwar2017pornography}, perform a combination of age estimation and pornography detection \cite{macedo2018benchmark} or synthetic data \cite{research1}. However, due to legal restrictions in maintaining a database of CSAM images, all current works are based on either unrealistic images \cite{research1}, or validated by authorities in small datasets \cite{vitorino2016two, macedo2018benchmark, gangwar2017pornography} that hardly represents the true data distribution in the internet \cite{ncmec_google}.

\textbf{Adversarially modified data samples } Adversarial inputs are intentionally crafted small perturbations to malicious inputs to elude detection from a model. For text applications, this can include injecting random noise that does not dramatically alter the understanding by a human.  Substitutions such as replacing "before" with "b4", homoglyph substitutions, and other substitutions, such as using "Lo7ita" instead of "Lolita" \cite{woodbridge2018homoglyph}. The effects of adversarial modifications in text classification have been explored for different NLP techniques, including classification \cite{agarwal2007much}, machine translation \cite{belinkov2017synthetic} and word embeddings \cite{heigold2017robust}. Depending on what kind of information is available to the adversary, it distorts portions of the text most likely to contain a signal important to the classification task.

\textbf{CSAM File Metadata Classification }
While significant efforts have focused on the images themselves, some researchers have looked for complementary signals to help CSAM identification. Such measures include queries that return CSAM in search engines, file metadata, and conversations that imply grooming or exchange of CSAM \cite{thorn_grooming}. Other efforts have used textual signals to identify where CSAM might be located,  such as keywords related to website content\cite{westlake}, using NLP analysis \cite{peersman2016icop,peersman2018detecting, al2020file}, conversations\cite{bogdanova2014exploring}. Our work falls into this category.
Previous works have found that perpetrators tend to use a specific CSAM vocabulary to name files \cite{peersman2016icop}. For this reason, using file paths, which is the combination of the file location and file name, is a promising approach for CSAM identification. To the best of our knowledge, there is one related work that aims to identify CSAM based solely on file path \cite{al2020file}. However, it does not address important questions such as classifiers robustness against adversarial examples and performance in out-of-sample benign datasets.

\section{Preliminaries}
In this section we provide an overview of the methods and algorithms utilized in our experiments.

\subsection{Text Preprocessing}

We present two different concepts utilized for text vectorization: term-frequency inverse-document-frequency (TF-IDF) and character-based quatization.

\textbf{TF-IDF} This techniques attributes weights to words (or sequences of characters) in a text \cite{TFIDF}. First it computes the term-frequency (TF), which is the number of times a term occurs in a given document. The inverse-document frequency-component (IDF) is computed as:
$$
\textnormal{IDF}(t) = \log \frac{1+n}{1+df(t)} + 1
$$
where $n$  is the total number of documents in the document set, and  $df(t)$ is the number of documents in the document set that contain term. For each term it is computed the product of the TF and IDF components. The resulting TF-IDF vectors are then normalized by the Euclidean norm.

\textbf{Character-Based Quantization} This type of text preprocessing is starts by defining an alphabet of size $m$ as the input language, quantized using 1-of-m encoding. Each textual input, of length $l$, is then transformed into a sequence of such $m$ sized vectors with fixed length $l$. Texts with more than $l$ characters are truncated and the exceeding initial characters are discarded. If the text is shorter than $l$, it is padded with zeroes on the left.Characters that are not in the alphabet are quantized as all-zero vectors. 

\subsection{Learning Algorithms}

We use several learning algorithms that have been successfully applied short text classification. We consider two broad categories of approaches: i) Traditional machine learning models, ii) and Neural networks models.

\subsubsection{Traditional Machine Learning}

\textbf{Logistic Regression }
This classification algorithm is a discriminative classifier that models the posterior probability $P(Y|X)$ of the class $Y$ given the input features $X$ by fitting a logistic curve to the relationship between $X$ and $Y$. Model outputs can be interpreted
as probabilities of the occurrence of a class \cite{Ng:2001}.

\textbf{Naive Bayes } Conditional probability model that assumes independence of features: given a problem instance to be classified, represented by a vector 
$\mathbf {x} =(x_{1},\dots ,x_{n})$
 representing some n features, it assigns to this instance probabilities 
$P(C_{k}\mid x_{1},\dots ,x_{n}) $ for each of K possible outcomes or classes $ C_{k}$. 
The problem with the above formulation is that if the number of features n is large or if a feature can take on a large number of values, then basing such a model on probability tables is infeasible. The model is reformulated to become more tractable. Using Bayes' theorem and assuming independence of the feature variable's, the conditional probability can be decomposed as: 
$$P(C_{k}\mid \mathbf {x} )={\frac {P(C_{k})\ P(\mathbf {x} \mid C_{k})}{P(\mathbf {x} )}}$$ 

\textbf{Boosted Decision Trees}
Model based on ensembles of trees, where each tree is trained using a boosting process in which each subsequent tree is built with weighted instances which were misclassified by the previous tree \cite{Freund:1997}.  Classification of a new instance with a trained ensemble of trees is based on a simple majority vote of the individual trees. 

\subsubsection{Neural Networks}

\textbf{Convolutional Neural Networks} 
One-dimensional Convolutional Neural Networks (CNNs) are a good fit when
the input is text, treated as a raw signal at the character level \cite{lecun}. CNN's automatically learn filters to detect patterns that
are important for prediction. The presence (or lack) of these
patterns is then used by the quintessential neural network
(multilayer perceptron) to make predictions. These
filters, (also called kernels) are learned during backpropagation. 

\textbf{Long Short-Term Memory network} This flexible architecture generalizes manual feature extraction via n-grams, for example, but instead learns dependencies of one or multiple characters, whether in succession or with arbitrary separation. The long short-term memory network (LSTM) layer can be thought of as an implicit feature extraction instead of explicit feature extraction (e.g., n-grams) used in other approaches. Rather than represent file paths explicitly as a bag of n-grams, for example, the LSTM learns patterns of tokens that maximize the performance of the second classification layer.

\section{Model Training}\label{classification}
In this section we describe the training dataset and a detailed description of our file path classifiers. 

\subsection{Training Dataset}
Our supervised learning approach to identify CSAM file paths utilizes a binary labeled dataset (CSA versus non-CSA). To separate the dataset into independent training and test sets, we split the data by storage system information (e.g., drive letter designations) do not leak information from the training to test set. Our dataset consists of real file paths collected by Project VIC International\footnote{https://www.projectvic.org}. The data consists of 1,010,000 file paths from 55,312 unique storage systems.  File paths are strings that contain location information of a file (folders) in a storage system and the file name. In Table \ref{dataset} we present details on the different types of content that constitute the dataset and the number of samples for each type of content.

\begin{table}[h]
\caption{Project VIC dataset set description. The dataset is used for model training and model testing. It contains non-pertinent (negative class) file paths and different types of file paths of child exploitive and child sexual abuse material (positive class).}
\begin{center}
\begin{tabular}{c c c}
\midrule
\textbf{Content Type} & \textbf{Label} & {\textbf{Samples}}  \\
\midrule
Non-pertinent &0  &  717,448\\
\\
Child Sexual Abuse Material (CSAM) & 1    & 33,901\\
\\
Child Exploitive/Age Difficult & 2   & 250,724\\
\\
CGI/Animation Child exploitive & 3     & 7,927\\
\midrule

\end{tabular}
\label{dataDescription}
\end{center}
\label{dataset}
\end{table}

We map the multiple classes provided by Project VIC into binary labels (CSAM vs. non-CSAM). Labels 1,2 and 3 are mapped to CSAM (292,552 file paths); label 0 is mapped to non-CSAM (717,448 file paths). 

\textbf{File Path Characteristics } The distribution of file path length help us define the size of the character embedding vectors in our deep neural networks models.

\begin{figure}
\centering
\includegraphics[width=3in]{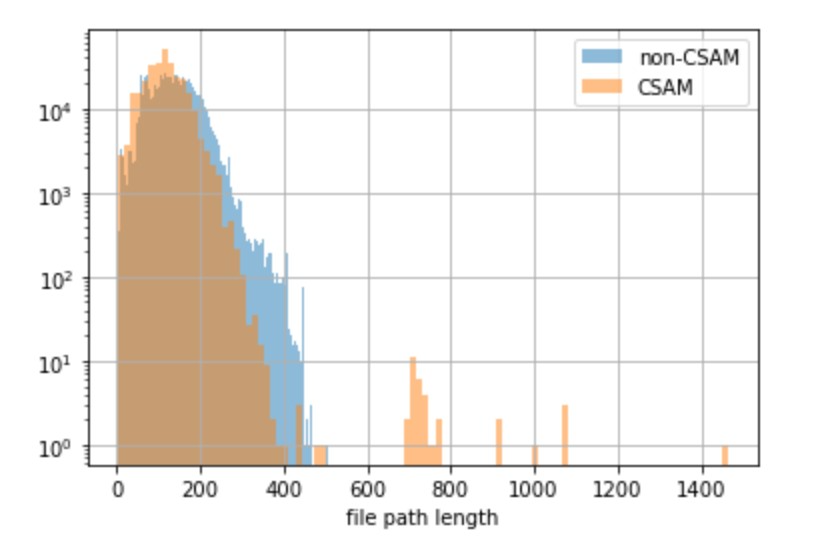}
\caption{The distribution of number of characters in file path dataset. Over 95\% of file paths are under 300 characters. Model input vector size is defined based on this distribution.}
\label{fig_length}
\end{figure}

Figure \ref{fig_length} shows the distribution of file path lengths in the dataset. Only 4,685 file paths have more than 300 characters. Our model takes as input vectors of at most 300 characters. For file paths with more than 300 characters, we truncate the file path by discarding the initial characters and keeping only the last 300 characters. For file paths with less than 300 characters, we pad with zeros on the left.

\begin{figure*}
\centering
\includegraphics[width=4.4in]{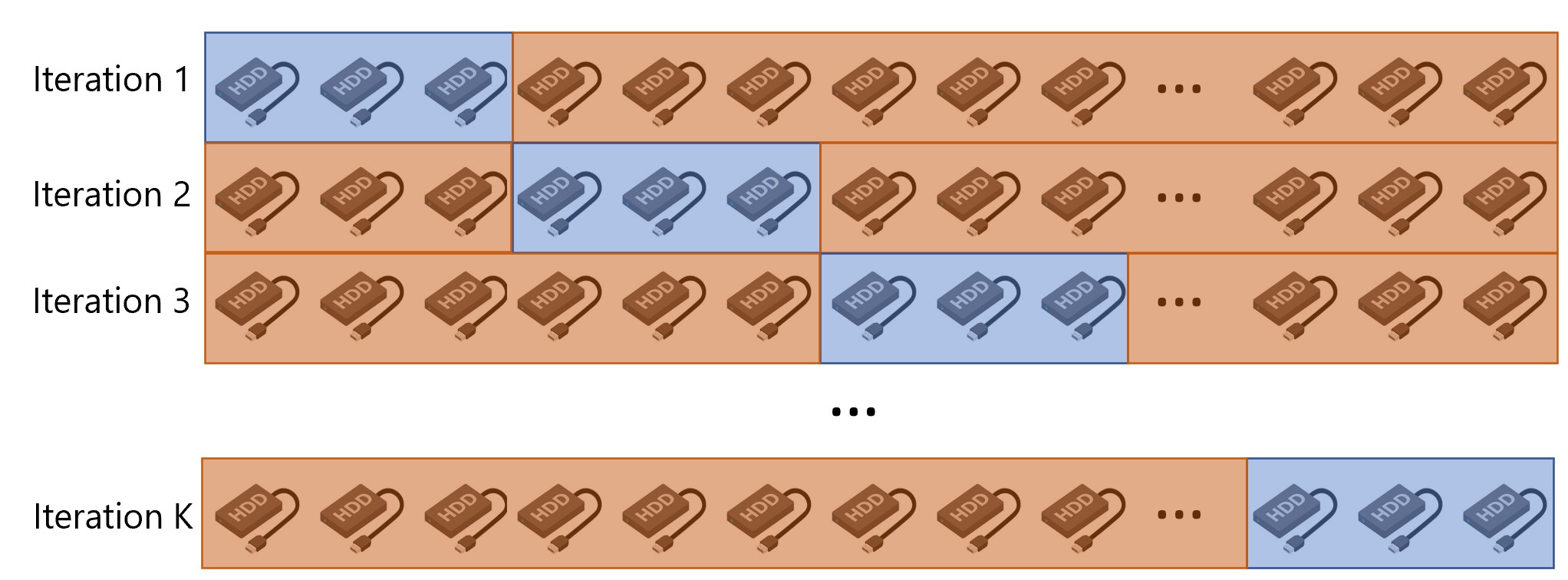}

\caption{Data split for K-fold cross validation. The data is partitioned by storage system ids. File paths from a same storage system are all assigned to a same data fold. The data folds in the orange shaded areas represent the training data in each iteration, and the data folds in the blue shaded areas represent the test data.}
\label{fig_sys}
\end{figure*}

\textbf{Cross Validation Data Split } We use a K-Fold Cross Validation methodology in our experiments with $K$=10. To guarantee the independence of file paths in the different partitions of the data, we create the random data folds by splitting the data by storage system information, as illustrated in Figure \ref{fig_sys}. 

The information before the first backlash of a file path specifies the external storage system or a laptop/desktop. This information is used to partition the dataset for cross-validation.

\subsection{File Path-Based CSAM Classifiers}

Our work investigates three approaches for CSAM file path classification. 

\begin{itemize}
    \item[1] \textbf{Bag-of-words } This approach encodes the file path string into a vector of words. The weights of the words are attributed using TF-IDF. We utilize the resulting vectors as input to traditional machine learning classifiers (logistic regression, boosted decision trees, and Naive Bayes).
    \item[2] \textbf{Character $N$-grams } A list of character sequences on size $N$ encodes the file path. The weights of the sequences are attributed using TF-IDF. The resulting vectors of the character sequences are used with traditional machine learning classifiers (logistic regression, boosted decision trees, and Naive Bayes);
    \item[3] \textbf{Character quantization } Sequences of encoded characters are used with a convolutional neural network (CNN) and a long short-term memory network (LSTM).  
\end{itemize}

\subsubsection{Implementation Details}\label{bow}
\hspace{5mm}\textbf{Bag-of-Words } For each file path, we consider a \textit{word} to be a sequence of alphanumeric characters that are separated by a dash, slash, colon, underscore or period. The bag-of-words model is constructed by selecting the
5,000 most frequent \textit{words} from the training subset. We utilize this text representation in combination with TF-IDF. The dataset of vectorized file paths is used as input to three different learning algorithms: logistic regression, naive Bayes and boosted decision trees.

\textbf{Bag-of-Ngrams } We extract from each file path string its n-grams, for $n \in \{1,2,3\}$. The set of n-grams of a string $s$ is the set of all substrings in $s$ of length n. We construct the bag-of-ngrams models by selecting the 50,000
most frequent n-grams (up to 3-grams) from the training dataset. We utilize this text representation in combination with TF-IDF. The dataset of vectorized file paths is used as input to three different learning algorithms: logistic regression, naive Bayes and boosted decision trees.

\textbf{Neural Networks } The alphabet used in all of our models consists of $m$ = 802 characters, including English letters, Japanese characters, Chinese characters, Korean characters, and special alphanumeric characters. The alphabet is the set of all unique characters in the training data.

All neural network architectures start with an embedding layer that represents each character by numerical vector. The embedding maps semantically similar characters to similar vectors, where the notion of similarity is automatically learned based on the classification task at hand. The variant of LSTM architecture used in our work the common "vanilla" architecture as used in \cite{woodbridge}.

\begin{figure}
\centering
\includegraphics[width=3.2in]{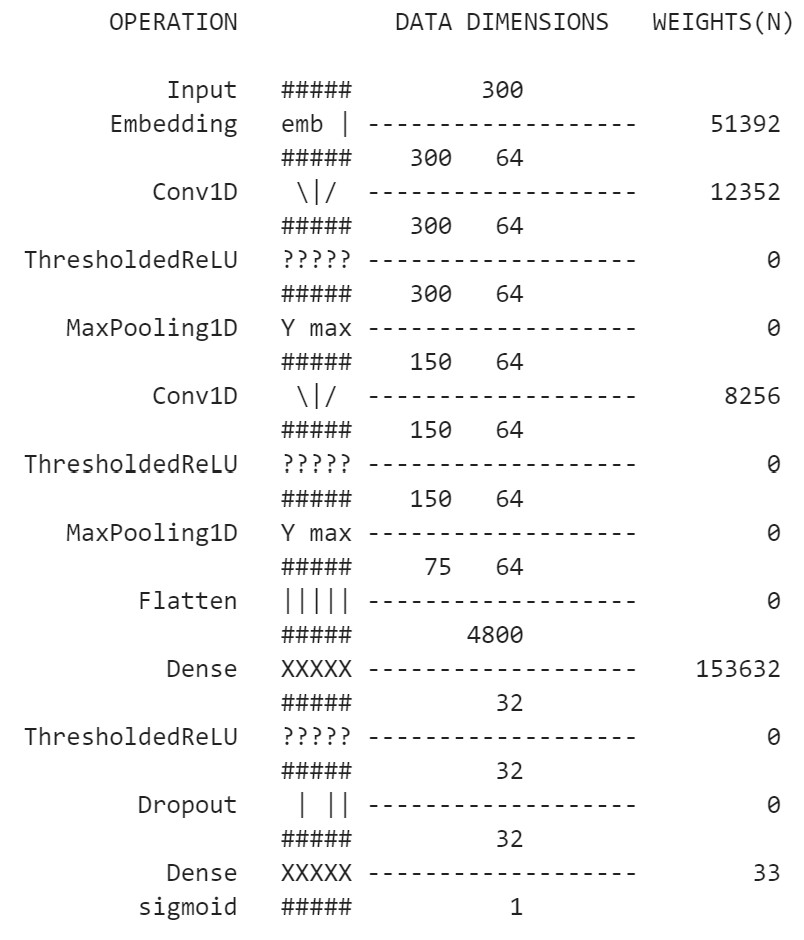}
\caption{Diagram of the deep neural network architecture with CNN layers used for training one of our charCNN model. All data dimensions and number of weights in each layer of our charCNN model are indicated in the above diagram.}
\label{archcnn}
\end{figure}

\begin{figure}
\centering
\includegraphics[width=3.2in]{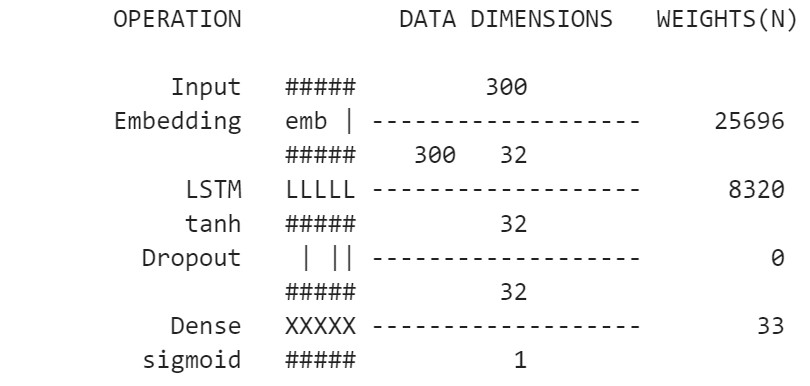}
\caption{Diagram of the deep neural network architecture with LSTM layer used for training one of our charLSTM model. All data dimensions and number of weights in each layer of our charLSTM model are indicated in the above diagram.}
\label{archlstm}
\end{figure}

Figures \ref{archcnn} and \ref{archlstm} show detailed information of both architectures(charCNN and charLSTM), including data dimensions and the number of weights in each layer. Through out this paper, we will refer to our CNN-based neural network and our LSTM-based neural network as charCNN and charLSTM, respectively.

\section{Model Evaluation I: Test Dataset from Project VIC }\label{section:evaluation}

\subsection{Evaluation Dataset}
We present our results for all our classifiers in Table \ref{TAB:Clean}. All performance metrics were measure using a 10-fold cross-validation on the Project VIC dataset. 

For each of our classifiers, we report the mean and the standard deviation over the folds for the area under the ROC curve (AUC), accuracy, precision, and recall for predicting CSAM files. We focus on two primary metrics for model comparison: Recall and AUC. Additionally, we assess all machine learning models' generalization by looking into the standard deviations over the cross-validation folds.

\subsection{Analysis}

\begin{table*}[t]
\caption{Model Evaluation I - Performance Metrics. Experiments with traditional machine learning and neural networks using Project VIC's dataset. We evaluate the AUC-ROC, accuracy, precision and recall. These results were measured across 10-folds in a cross-validation setting. For each metric, we report the mean ($\mu$), and the standard deviation ($\sigma$).}
\begin{center}
\begin{tabular}{ l l cc cc cc cc }
\midrule
\textbf{Features}& \textbf{Algorithm}& \multicolumn{2}{c}{\bf AUC}& \multicolumn{2}{c}{\bf Accuracy}& \multicolumn{2}{c}{\bf Precision} & \multicolumn{2}{c}{\bf Recall} \\
\\
& & $\mu$ & $\sigma$ & $\mu$ & $\sigma$ & $\mu$ & $\sigma$ & $\mu$ & $\sigma$\\

\midrule
Bag of Words &   Logistic Regression  & 0.967 &	0.035 &  0.922 & 0.062  & 0.904 &	0.090 &	0.787 &	0.202\\
\\
Bag of Words      &      Naive Bayes &0.972 &	0.011& 0.927&	0.032&	0.875&	0.070&	0.859&	0.114 \\
\\
Bag of Words     &  Boosted Trees  & 0.982&	0.013& 0.934&	0.062&	0.903&	0.096&	0.827&	0.203 \\

\midrule 
Bag of N-grams      &  Logistic Regression  & 0.980&	0.021& 0.931&	0.060&	0.919&	0.088&	0.793&	0.202 \\
\\ Bag of N-grams      &  Naive Bayes  & 0.958&	0.023& 0.929&	0.032&	0.839&	0.083&	0.913&	0.085 \\
\\ Bag of N-grams      &  Boosted Trees  & 0.983&	0.015& 0.931&	0.060&	0.906&	0.094&	0.822& 0.203 \\

\midrule Character quantization      &  CNN  & \textbf{0.990}&	0.011& \textbf{0.968}&	0.019&	\textbf{0.938}&	0.034&	\textbf{0.943}&	0.060 \\
\\ Character quantization      &  LSTM  &  0.977&	0.029& 0.930&	0.072&	0.862&	0.144&	0.846&	0.213 \\

\hline
\end{tabular}
\label{TAB:Clean}
\end{center}
\end{table*}

\textbf{Traditional Machine Learning Models} There are significant advantages of traditional machine learning models in comparison with deep neural networks. Understanding how well these models perform can help scientists and investigators leverage such models' most remarkable characteristics: feature interpretability. The most relevant predictive tokens, or n-grams, can give clues about vocabulary words in the dataset and utilize it in other CSAM detection systems. In table \ref{TAB:Clean}, we observe that the model trained with bag-of-words and bag-of-ngrams operates in similar AUC and accuracy ranges. When analyzing recall rates of traditional models, we note that both naive Bayes models have the highest average rates and lowest standard deviations.
The naive Bayes with bag-of-ngrams features presents the best recall of all traditional models, of about 0.91. Among the other models trained using bag-of-ngrams, naive Bayes presents a much smaller recall standard deviation ($\sigma$ = 0.085) when compared to logistic regression ($\sigma$ = 0.20) and boosted decision trees($\sigma$ = 0.20).

Although the evaluation of CSAM classification models heavily relies on recall rates, when deploying a model in an environment that potentially analyzes hundreds on thousands of file systems and consequently millions of file paths, precision can become the most significant metric. The burden of having several thousands of false positives can result in an inefficient process and potentially delay investigations and discovery of true positives. The AUC metric captures the ability of a classifier to operate with high recall when low false positive rates are necessary.  By analyzing the traditional models' AUC, we observe that boosted decision trees overall performs better than the two other techniques.

\textbf{Neural Networks Models} We were able to achieve the best performance across all categories with deep neural network architecture. We trained two different architectures: a layered CNN and an LSTM. The LSTM model achieves results very similar to the bag-of-words naive Bayes classifier. However, our CNN model consistently outperforms all the other models, both in mean performance metrics across all folds and in the lowest standard deviation. The recall of over 0.94 and precision over 0.93 makes this model an excellent candidate as an investigative tool in environments with large volumes of storage systems.

\section{Model Evaluation II: Test Dataset with Adversarial Inputs}
In classical machine learning applications, we assume that the underlying data distribution is stationary at test time. However, a testing pipeline of models aimed at detection of illegal activities should anticipate an intelligent, adaptive adversary actively manipulating data. We know perpetrators purposely add typos and modifications to file identifiers \cite{peersman2016icop} to evade blocklists and machine learning-based detection mechanisms. We modify our test dataset to simulate an adversary actively changing the file paths to elude the classifiers.

In our testing scenario, the adversary can send file paths to the model, but it does not have access to model outputs. The only allowed action to the adversary is to make changes to the file paths. 

Although the adversary is willing to change the file paths to evade detection, it is not interested in completely changing the file path's words (and meaning). The files are often shared among perpetrators, and the file name is usually used to identify file contents. Therefore, the adversary wants to make the maximum amount of changes without compromising the human comprehension of the meaning of the string of characters. The number of modifications in the file path is called the adversary's budget.



\begin{definition}{Adversarial Examples.} Given a model $\mathcal{F: X \rightarrow Y}$, which maps the input space $\mathcal{X}$ to the set of labels $\mathcal{Y}$, a valid adversarial example $x_{adv}$ is generated by altering the original data example $x \in \mathcal{X}$ and should conform to the following requirement: $S(x_{adv, x}) \leq \epsilon$, where $S: \mathcal{X \times X}\rightarrow \mathcal{R}^{+}$ is a similarity function and $\epsilon \in \mathcal{R}^{+}$ is a constant modeling the budget available to the adversary, i.e. the allowable \textit{size} of a modification.
\end{definition}

\textbf{Threat Model } In our threat model, the attacker is not aware of the model architecture, parameters and does not have access to the confidence scores predicted by the model. The attacker attempts to cause an integrity violation in the model by modifying the input under bounded perturbation size: a hard-label black-box evasion scenario \cite{goodfellow2014explaining}.  The only knowledge the adversary has about the model is the input space $\mathcal{X}$ and the output space $\mathcal{Y}$.

\subsection{Evaluation Dataset}
\begin{figure*}
\centering
\includegraphics[width=6.5in]{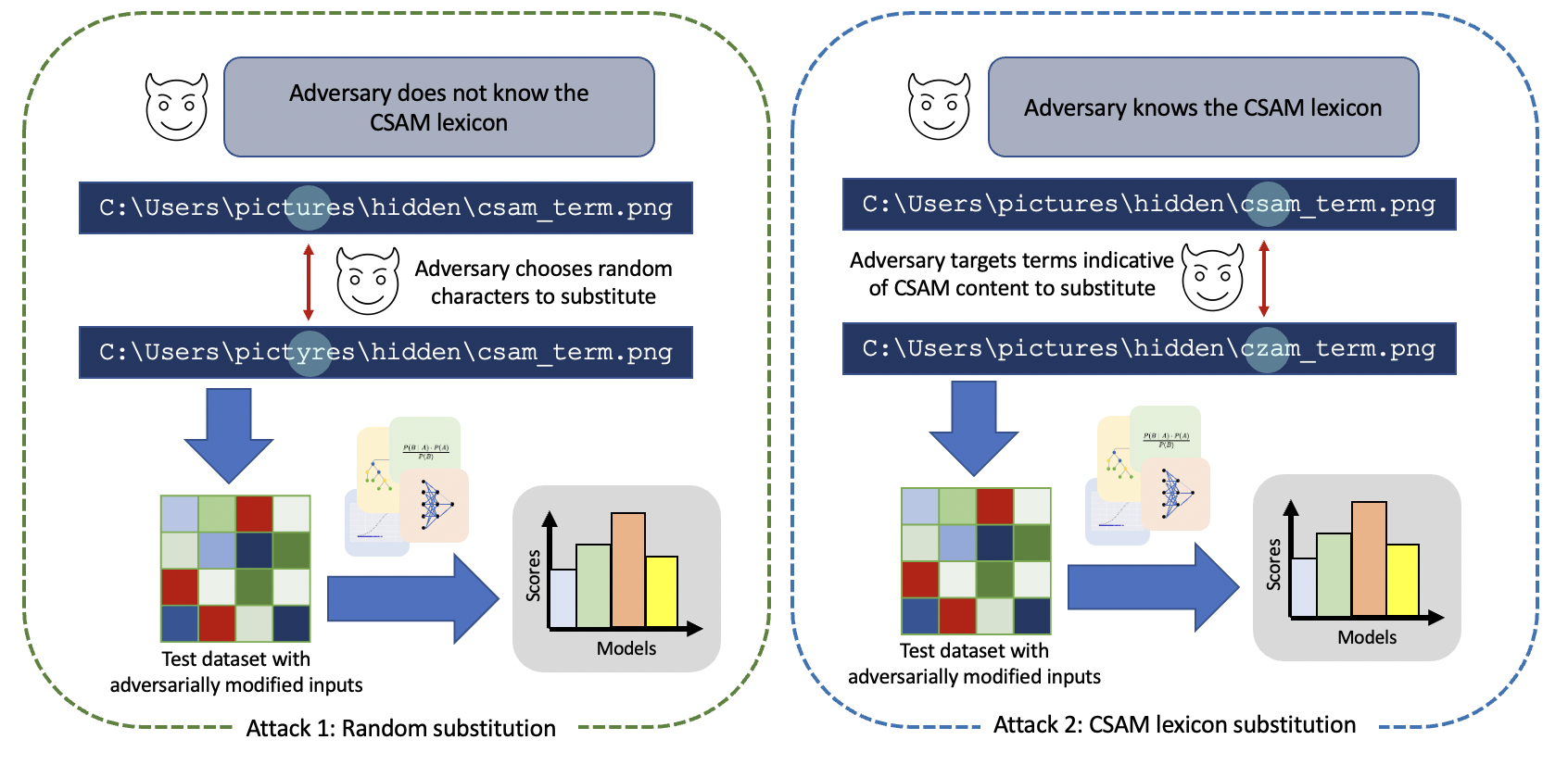}
\caption{Adversarial Inputs Generation. We generate adversarial inputs based on two different adversarial attacks. (1) In the random substitution attack, the adversary chooses a random position in the file path string and substitute the character in the selected position. (2) The CSAM lexicon substitution attack allows the adversary access to a CSAM lexicon. The adversary substitutes characters in terms that are present in the CSAM lexicon.}
\label{fig_attack}
\end{figure*}

In our experiments we modify the test dataset based on two variations of the threat model, shown in Figure \ref{fig_attack}. 


\textbf{Attack 1: Random Substitutions } The adversarial examples are generated by randomly selecting a position in the file path string and substituting the character in the selected position by a random alphanumeric character. This technique has been previously used to attack language models \cite{belinkov2017synthetic}. We evaluate our models for an adversarial budget of 10\%, 15\%, and 20\% of file path length. 

\textbf{Attack 2: CSAM Lexicon Substitution } We assume the adversary has access to a CSAM lexicon, a list of words used by perpetrators to mark media containing CSAM. The adversary uses the CSAM lexicon to identify words in the file path that indicates the presence of CSAM. As illustrated in Figure \ref{fig_attack}, in this attack, the adversary first locates a word from the lexicon in the file path and then substitutes characters.

The file path modification occurs as follows: The adversary verifies which words in the CSAM lexicon are present in the file path. For every word, the adversary randomly modifies one character by randomly selecting a position in the word string and substituting the character in the chosen position by a random character. The number of allowed word replacements is determined by the budget $\epsilon$.

For our experiments, we create the CSAM lexicon list using Odds Ratio, a widely used technique in information retrieval, and used for feature selection and interpretation of text classification models \cite{mladenic1998feature}.

First, we identify which tokens are more likely to appear in CSAM file paths. We extract all tokens from the dataset of file paths as described in section \ref{bow}. We calculate the odds of the keyword being part of a CSAM file path and the odds of the keyword being part of a non-CSAM file path for all keywords. The Odds ratio of a word $w$ is computed as: 
$$
\textnormal{Odds Ratio} = \frac{\textnormal{odds of $w$ appear in CSAM file}}{\text{odds of $w$ appear in non-CSAM}}
$$
The CSAM lexicon comprises all keywords with Odds Ratio greater than two. We make this list available to the adversary.


The adversarial modifications are done in the test fold of the 10-fold cross-validation on the Project VIC dataset.

\subsection{Analysis} 

We evaluate the impact of adversarial modifications in test samples on the model's performance. We are interested in i) understanding which machine learning techniques are more robust when the data is adversarially modified at test time, ii) and how much the performance of the models change. All attacks target only CSAM file paths, and therefore we only evaluate the variation in recall rates. Additionally, we analyze the mean deviation in confidence scores for all models.

\textbf{Random Substitutions } Under this scenario, an adversary randomly modifies a percentage of the file path by randomly selecting characters and replacing them with random characters. A reasonable adversary budget in this scenario is between 10 \% and 15\%. Previous works have also considered this same percentage range for perturbing text strings \cite{textfool}. Considering that most file paths have a length between 40 and 200 characters, this results in changing between 6 and 30 characters in each file path. To stress-test our models, we also analyze the performance of our models under a 20\% change. 

\begin{figure}
\centering
\includegraphics[width=3.4in]{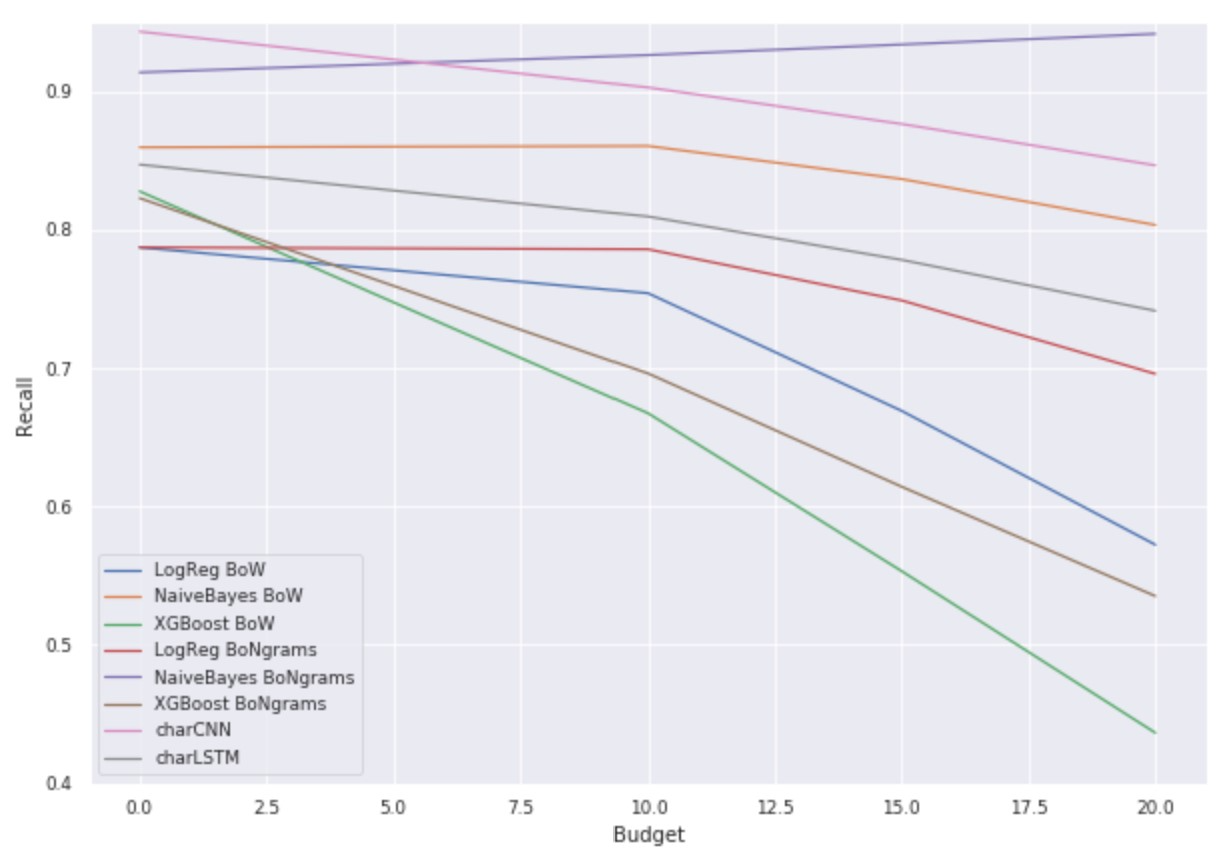}
\caption{Model Evaluation II - recall variation as adversary budget increases in a random substitution attack. Our best performing model, a charCNN, suffers a decrease in recall as the adversary budget increase, yet the recall is still over 0.80 even when the 20\% of the characters of a CSAM file path is modified. Surprisingly, the model trained using Naive bayes (with Bag-of-Words as the text preprocessing technique) suffers an increase in recall as the adversary budget increase.}
\label{fig_rbb}
\end{figure}

Table \ref{TAB:attack} shows details on the confidence score variations for different adversarial budgets.

 \begin{table}
\caption{Model Evaluation II - model confidence decrease under adversarial inputs. Experiment with random substitution attack, for adversarial budget of  10\%, 15\% and 20\%. We evaluate the mean decrease in model cofidence ($\mu$) and the model confidence decrease variance ($\sigma$) for traditional machine learning algorithms - Logistic Regression (LR), Naive Bayes (NB), and Boosted Trees (BT) and neural networks - Convolutional Neural Networks (CNN) and Long Short-Term Memory Network (LSTM). The text preprocessing methods used use traditional machine learning algorithms are bag-of-words (BW) and bag-of-n-grams (BN). The text preprocessing method used with neural networks algorithms is character quantization (CQ).}
\begin{center}
\begin{tabular}{l l cc cc cc }
\hline
\textbf{Feature}& \textbf{Model}& \multicolumn{2}{c }{\bf $\epsilon$ = 10\%}& \multicolumn{2}{c }{\bf $\epsilon$ = 15\%}& \multicolumn{2}{c }{\bf $\epsilon$ = 20\%}\\
\\
& & $\mu$ & $\sigma$ & $\mu$ & $\sigma$  & $\mu$ & $\sigma$ \\
\midrule
BW &   LR  &  
0.05	& 0.19 & 0.09 &	0.24 &
0.13 &	0.28 
\\
\\
BW      &      NB &  $\approx$0	& 0.25 & 0.03	& 0.29 &	
0.07 &	0.32 
\\
\\ 
BW    &  BT  &  0.16 &	0.35 & 0.26 &	0.40 &	
0.35 &	0.40 
\\

\midrule 
BN     &  LR & 0.05 &	0.11 & 0.08 &	0.13 &	
0.12 &	0.15 
\\
\\
BN      &  NB  & -0.01 &	0.18 &	-0.01 &	0.2 &	-0.02 &	0.21 \\
\\ BN &  BT  & 0.12 &	0.30 & 0.19 &	0.35 &	
0.26 &	0.38 
\\

\midrule CQ      &  CNN  & 0.03 &	0.13 & 0.05 &	0.18 &	
0.08 &	0.22
\\
\\ CQ     &  LSTM  & 0.05 &	0.22 & 0.09 &	0.27 &	
0.13 &	0.31

\\

\hline
\end{tabular}
\end{center}
\label{TAB:attack}
\end{table}

Figure \ref{fig_rbb} demonstrates the variation in recall rates as the percentage of random flipped characters increases. For an adversarial budget of 15\%, we observe a decrease in recall rates of 0.02 for bag-of-words and naive Bayes and 0.07 in the CNN model. Interestingly, bag-of-ngrams naive Bayes suffers an increase in recall rates after flipping a percentage of the characters in the file path. This phenomenon results from the fact that modifications in the file paths can also increase the model's confidence score output. The boosted decision trees models undergo the most significant decrease in recall rates. A possible model overfitting can explain this to this specific dataset distribution. When flipping 20\% of characters, the deep neural networks models' recall rates decrease $\approx 0.1$, and bag-of-words naive Bayes decreases $\approx 0.05$, and bag-of-ngrams naive Bayes, once again presents an increase in its recall rate.

\begin{table}
\caption{Model Evaluation II - model confidence decrease under adversarial inputs. Experiment with CSAM lexicon substitution attack, for adversarial budget of 1, 2, 3 and 4 characters. We evaluate the mean decrease in model cofidence ($\mu$) and the model confidence decrease variance ($\sigma$) for traditional machine learning algorithms - Logistic Regression (LR), Naive Bayes (NB), and Boosted Trees (BT) and neural networks - Convolutional Neural Networks (CNN) and Long Short-Term Memory Network (LSTM). The text preprocessing methods used use traditional machine learning algorithms are bag-of-words (BW) and bag-of-n-grams (BN). The text preprocessing method used with neural networks algorithms is character quantization (CQ).}
\begin{center}
\begin{tabular}{l l cc cc }
\hline
\textbf{Feature}& \textbf{Model}& \multicolumn{2}{c}{\bf $\epsilon$ =  1}& \multicolumn{2}{c}{\bf $\epsilon$ = 2} \\
& & $\mu$ & $\sigma$ & $\mu$ & $\sigma$ \\
\hline
BW &   LR  &  
0.01 &	0.03 &
0.02 &	0.05 
\\
\\
BW      &      NB &
$\approx$0 &	0.05 &
$\approx$0 &	0.06 
\\
\\ 
BW     &  BT  & 
0.01 &	0.06&
	0.02&	0.07
\\
\midrule BN     &  LR  & 
0.01 &	0.01 &
0.01 &	0.02 
  \\
\\ BN     &  NB  & 
$\approx$0 &	0.05 &
$\approx$0 &	0.06 
 \\
\\ BN     &  BT  & 
$\approx$0 &	0.05 &
0.01 &	0.06 
\\
\midrule CQ     &  CNN  & 
$\approx$0 &	0.02 &
$\approx$0 &	0.03 
 \\
\\ CQ    &  LSTM  & 
$\approx$0 &	0.04 &
$\approx$0 &	0.05 
\\

\midrule
\textbf{Feature}& \textbf{Model} & \multicolumn{2}{c}{\bf $\epsilon$ = 3} & \multicolumn{2}{c}{\bf $\epsilon$ = 4} \\
& & $\mu$ & $\sigma$ & $\mu$ & $\sigma$ \\
\hline
BW &   LR  &  

0.03 &	0.05 &
0.03 &	0.06
\\
\\
BW      &      NB &
$\approx$0 &	0.07 &
$\approx$0 &	0.07
\\
\\ 
BW     &  BT  & 

	0.02&	0.08&
	0.03&	0.09
\\
\midrule BN     &  LR  & 

0.02 &	0.02 &
0.02 &	0.02
  \\
\\ BN     &  NB  & 

$\approx$0 &	0.06 &
$\approx$0 &	0.07
 \\
\\ BN     &  BT  & 

0.01 &	0.07 &
0.01 &	0.07\\
\midrule CQ     &  CNN  & 
$\approx$0 &	0.03 &
$\approx$0 &	0.04
 \\
\\ CQ    &  LSTM  & 
$\approx$0 &	0.05 &
$\approx$0 &	0.06\\

\midrule
\end{tabular}
\label{TAB:attack2}
\end{center}
\end{table}

Figure \ref{fig_rbb} demonstrates the variation in recall rates as the percentage of random flipped characters increases. For an adversarial budget of 15\%, we observe a decrease in recall rates of 0.02 for bag-of-words and naive Bayes and 0.07 in the CNN model. Interestingly, bag-of-ngrams naive Bayes suffers an increase in recall rates after flipping a percentage of the characters in the file path. This phenomenon results from the fact that modifications in the file paths can also increase the model's confidence score output. The boosted decision trees models undergo the most significant decrease in recall rates. A possible model overfitting can explain this to this specific dataset distribution. When flipping 20\% of characters, the deep neural networks models' recall rates decrease $\approx 0.1$, and bag-of-words naive Bayes decreases $\approx 0.05$, and bag-of-ngrams naive Bayes, once again presents an increase in its recall rate.

\begin{figure}[t]
\centering
\includegraphics[width=3.6in]{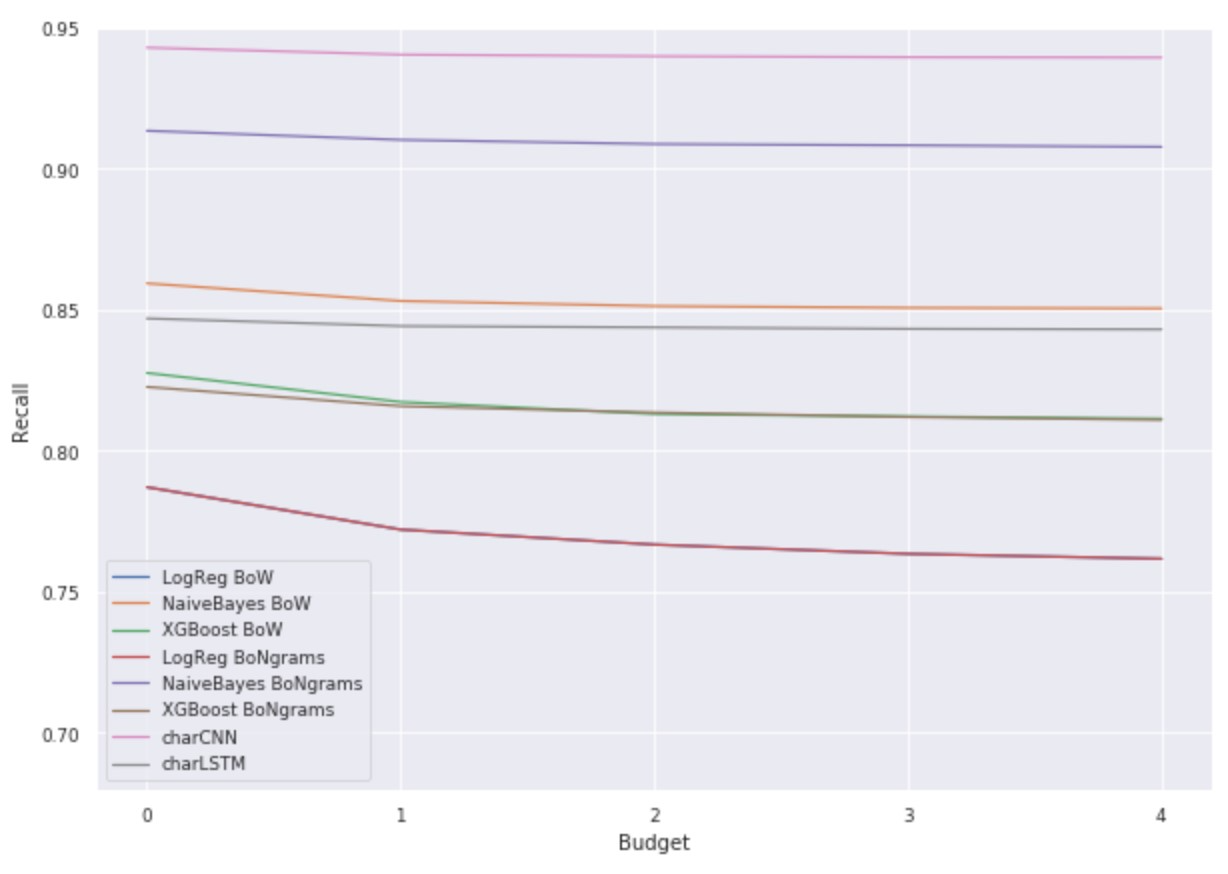}
\caption{Model Evaluation II - recall variation as adversary budget increases in a CSAM lexicon substitution attack. Our best performing model, a charCNN, does not suffer any significant recall decreases. The model trained using Logistic Regression (together Bag-of-N-grams) suffers the greater decrease in recall of all models.}
\label{fig_rbw}
\end{figure}

\textbf{CSAM Lexicon Substitution} Having access to a list of terms highly correlated with CSAM file paths, permits the adversary to make targeted changes in the CSAM file paths. This experiment allows the adversary to change one character per keyword, up to 4 keywords per file path. Figure \ref{fig_rbw} illustrates the recall variation as a function of the adversarial budget

As indicated in figure \ref{fig_rbw}, the most significant drop in accuracy happens for budget = 1, which can be justified by the fact that the adversary also has access to the Odds Ratio for each keyword. For budget = 1, the adversary will modify the keyword with the largest Odds Ratio; for budget = 2, the adversary will modify the keywords with the two largest Odds Ratio, and so on.

Overall, the targeted changes in the file paths result in small changes in the recall rates. Logistic regression and boosted decision trees have more considerable recall variations than naive Bayes and deep neural networks models. 

Confidence score variation details are described in table \ref{TAB:attack2}. It is easy to observe that the mean change in confidence score is less than 0.1 for all models and all budgets. However, we see examples in our experiments where a single change resulted in a drastic drop in the model confidence. 

Without access to the model output, we conclude that it is hard to craft an adversarial example close to the original sample, even when an adversary has access to a list of keywords highly correlated with the positive class.

\section{Model Evaluation III: Test with Benign Dataset from Open Data}

\begin{figure*}
\centering
\includegraphics[width=6in]{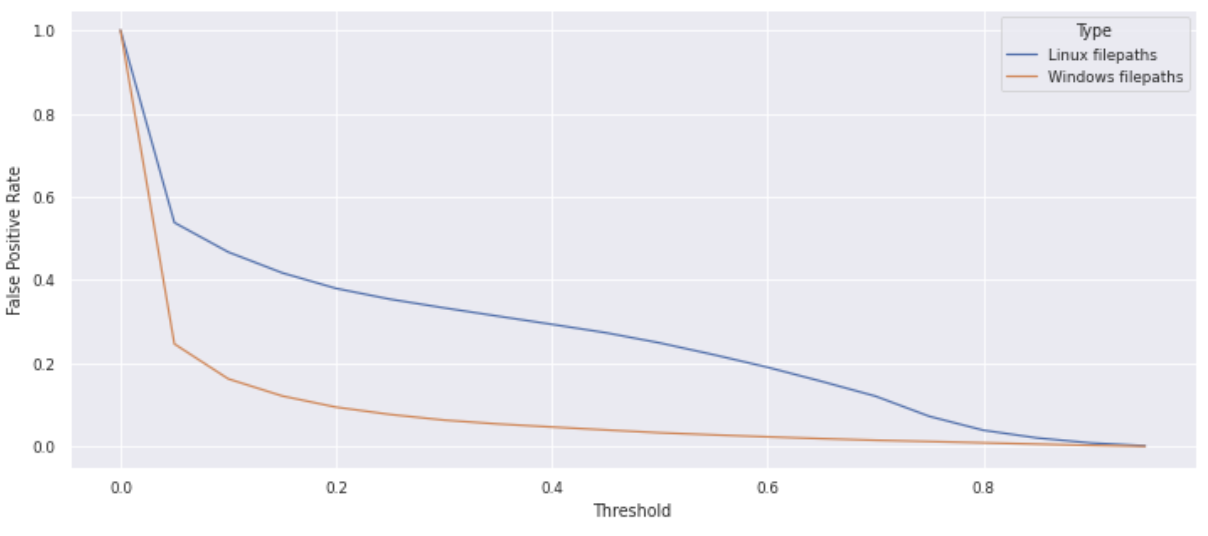}
\caption{Model Evaluation III - false positive rate as model confidence threshold changes. Our best performing model, a CNN-based model, exhibits very low false positive rates when used to score Windows file paths, with an FPR of 0.03 for a confidence threshold of 0.5, but achieves an FPR of 0.005 for confidence threshold of 0.85. The model presents a higher FPR on Linux file paths, where at a confidence level of 0.5 it exhibits an FPR of 0.24. However, it drops significantly for a higher confidence threshold, achieving an FPR of 0.02 at a confidence level of 0.85.}
\label{fig_fpr}
\end{figure*}

 A common measure of success for detection systems based on machine learning is the true positive rate, also know as recall.
 
 To guarantee a successful deployment the detection system, another metric of extreme importance is the false positive rate. The usual training pipeline evaluates this metric in data distributions that are similar to the training data. However, when possible, an extremely valuable experiment is to collect data from different distributions to understand model performance under different conditions. When a positive detection occurs, detection systems usually triggers an action, which is usually human data review. If the false positive rate is not well understood, and model operation correctly calibrate, it can cause a burden to content moderators, and jeopardize the deployment of the system.

\subsection{Evaluation Dataset}
The benign samples dataset was constructed using the publicly available common crawl dataset. We collected data from Common Crawl index CC-MAIN-2021-10. The WARC files utilized to construct our dataset are:

\begin{itemize}
    \item Linux file paths: we parsed the first 200 WARCs (00000-00199, inclusive) which resulted in over 73k unique paths.
    \item Windows file paths: we parsed 11821 WARCS (00000-12000, inclusive) which resulted in 32K unique paths.
\end{itemize}

We parsed the raw HTML, treating it as a Latin-encoded string.In each HTML, regular expression functions for identifying windows and linux filepaths are the following:

\begin{verbatim}
    windows_filepath_with_ext = r"([a-z]:\\
    ([a-z0-9() ]*\\)*[a-z0-9()]*\.(jpg|jpeg|
    png|gif|mp4|mov|m4a|m4v|mpg|mpeg|wmv|avi|
    flv|3gp|3gpp|3g2|3gp2|doc|docx|xls|xlsx|
    ppt|pptx|pdf))"
\end{verbatim}
\begin{verbatim}
    linux_filepath_with_ext = r"(/([a-zA-Z0-9
    ()]*/)*[a-zA-Z0-9()]*\.(jpg|jpeg|png|gif|
    mp4|mov|m4a|m4v|mpg|mpeg|wmv|avi|flv|3gp|
    3gpp|3g2|3gp2|doc|docx|xls|xlsx|ppt|pptx|
    pdf))"
\end{verbatim}

  After collecting the dataset using the functions above, we filtered Windows filenames to exclude “:\textbackslash u002F”. In Linux filenames we only keep the filepaths that begin with: /usr/, /home, /etc, /tmp, and  /var.

\subsection{Analysis}

The evaluation of model performance in independent datasets is essential to understand model generalization. Most importantly, CSAM file paths account for a small fraction of file paths in sharing platforms. We test our best-performing model against a dataset containing only benign file paths. We measured the false positive rate for the best-performing model, char-CNN, at different confidence thresholds.

As we can observe from figure \ref{fig_fpr}, for decision thresholds above 0.8, the false positive rate is low. For example, for Linux file paths, a decision threshold of 0.8 results in an FPR of $\approx 0.03$, whereas a decision threshold of 0.95 results in an FPR of $\approx 0.001$. For Windows file paths, a threshold of 0.8 prompts an FPR less than 0.01, while a decision threshold of 0.95 leads to an FPR less than 0.001. High decision thresholds are common design choices in detection systems, where only the high confidence samples are flagged and sent for human review.

Given the high AUC values for the character-based CNN, measured with Project VIC's dataset, it seems reasonable to believe that we can achieve a good recall for thresholds above 0.8 and thus guaranteeing a low false-positive rate. 

\section{Conclusion}

In this paper, we proposed a training and evaluation framework to develop deployment-ready machine learning models for CSAM file detection. Our framework includes a model training pipeline that covers several techniques for text representation and machine learning. Our evaluation pipeline covers real-world scenarios that surface when deploying a machine learning model for CSAM detection.  All testing scenarios are rigorously defined and easily reproducible. 

The proposed system for CSAM identification based solely on file paths has the advantage of not working directly with CSA photos or videos. The classifier is a medium agnostic CSAM detector of easy maintenance and reduced legal restrictions for acquiring training data. Our classifier achieves precision and recall rates over 0.90 in out-of-sample hard drives.  Our experiments also show that our models generalize well to identifying CSAM content in file storage systems and preserve low FPR in out-of-sample negative samples. Additionally, we present a testing framework to evaluate model robustness to adversarial attacks introduced at test time. 

The proposed framework is an essential addition to the available tools for CSAM detection. The community can leverage the proposed framework to train and evaluate models for CSAM metadata and short-text classification tasks, such as file path classification and CSAM search terms classification. 

In combination with PhotoDNA hash, computer vision tools, and other forensics tools, our CSAM file path classifier integrates a global toolset that enables organizations to fight the distribution of CSAM. 

Online child sexual abuse imagery falls into a category of content that should not be distributed or be present in file storage systems. The distributed nature of the internet makes CSAM detection a complex problem to solve. Automated tools and machine learning-based systems can help technology companies and investigation agencies rapidly identify such content and take the appropriate actions.

\ifCLASSOPTIONcaptionsoff
  \newpage
\fi



%
\bibliographystyle{plain}
\bibliography{csam}

%








\end{document}